\documentclass[letterpaper, 10 pt, conference]{ieeeconf}  
\IEEEoverridecommandlockouts                              

\usepackage{amsmath} % assumes amsmath package installed
\usepackage{graphicx} % for including graphics
\usepackage{amssymb}
\usepackage{booktabs, tabularx}
\usepackage{cite}
\usepackage{caption}
\usepackage{capt-of}
\usepackage{color}
 \usepackage{bm}

\begin{document}

\title{
% Visuomotor Diffusion Policy for Coordinated Mobile Dual-Arm Door Manipulation
Diffusion Policy for Coordinated Control of a Nonholonomic Mobile Base and Dual Arms in Door Opening and Passing
}

% $^{1}$
% Add your name
% \author{
%     Shangqun Yu, Sangjun Park, Matthew En, and Donghyun Kim \\
%     {\medium AlphaZ}
% \thanks{*}% <-this % stops a space
% \thanks{$^{1}$College of Information and Computer Sciences, University of Massachusetts Amherst, MA, U.S.
%         {\tt\small}}%
% }
% \author{Author Names Omitted for Anonymous Review. Paper-ID [add your ID here]}
\author{Shangqun Yu$^{1}$, Matthew En$^{1}$, Daniel Wu, Sangjun Park$^{1}$, Ziyi Zhou \\
Seyed Fakoorian$^{2}$, Donghyun Kim$^{1}$%
\thanks{$^{1}$Manning College of Information and Computer Sciences, University of Massachusetts Amherst. ({\tt\small donghyunkim@cs.umass.edu})}%
\thanks{$^{2}$alpha-z.}%
}

\maketitle

% \twocolumn[{%
% \renewcommand\twocolumn[1][]{#1}%
% \begin{center} 
%     \centering
%     \captionsetup{type=figure}
%     \includegraphics[width=\linewidth]{fig/hdsf.pdf}
%     \captionof{figure}{{\bf Diffusion-based Door Opening and Traversal Policy.} We trained a diffusion policy that enables a mobile manipulator to open and traverse a damped pull door, a task requiring tight coordination of perception, dual-arm manipulation, and base navigation. The learned policy executes the full long-horizon sequence of reaching, twisting, pulling, and passing, while also demonstrating robustness to external disturbances by detecting and recovering from them—an ability that is crucial for real-world deployment.}  
%     \label{fig:hcd}
% \end{center}%
% }]

\begin{figure*} [ht!]
    \centering
    \includegraphics[width=\linewidth]{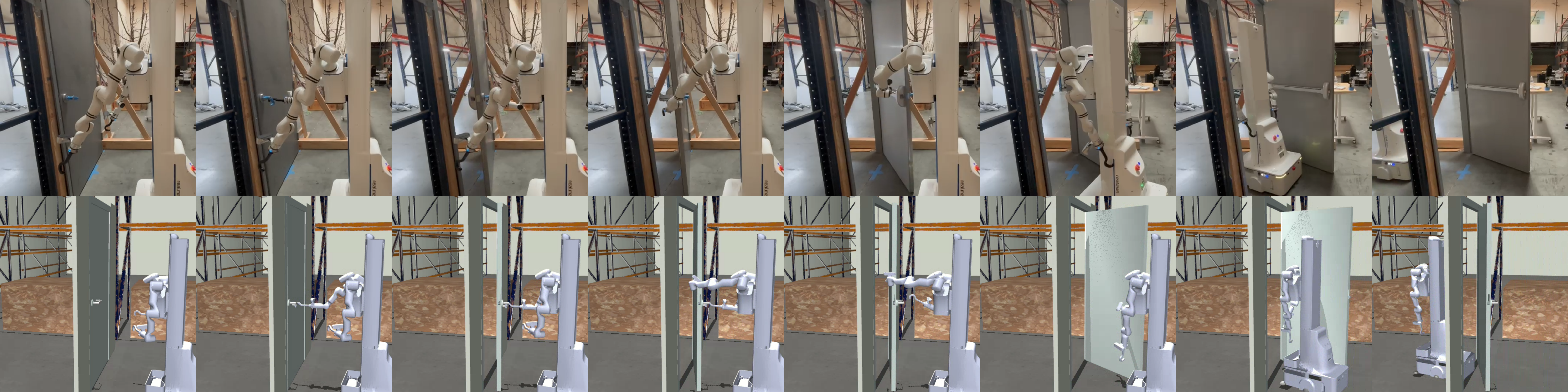}
    \captionof{figure}{{\bf Diffusion-based Door Opening and Traversal Policy.} We trained a diffusion policy that enables a mobile manipulator to open and traverse a damped pull door, a task requiring tight coordination of perception, dual-arm manipulation, and base navigation. The learned policy executes the full long-horizon sequence of reaching, twisting, pulling, and passing, while also demonstrating robustness to external disturbances by detecting and recovering from them—an ability that is crucial for real-world deployment.}  
    \label{fig:hcd}
\end{figure*}

%%%%%%%%%%%%%%%%%%%%%%%%%%%%%%%%%%%%%%%%%%%%%%%%%%%%%%%%%%%%%%%%%%%%%%%%%%%%%%%%
\begin{abstract}
Opening heavy, self-closing doors---especially those that require pulling---remains a long-standing challenge in robotics. Humans naturally employ both arms in a dexterous manner---rotating the handle, widening the gap, holding the door, switching arms when needed, and moving through while maintaining clearance. To replicate such behaviors, a robot must perform a long sequence of motions spanning multiple stages and interactions with different parts of the door. Traditional approaches rely on state machines that transition between manually defined stages (e.g., pulling after the knob is rotated, passing after the gap is sufficiently wide). While intuitive, these methods lack robustness, as hand-crafted trajectories fail to generalize to the diversity of real-world conditions without extensive engineering effort. Recent advances in imitation learning offer a scalable alternative, yet no existing visual-action model has demonstrated simultaneous coordination of a nonholonomic base and dual arms for the complete door opening and passing task.
In this paper, we tackle this complex, highly constrained problem using a diffusion-based visuomotor control policy. Our results demonstrate that a single end-to-end policy can be learned to execute long-horizon tasks requiring tight coordination between manipulation and locomotion. The resulting policy not only achieves a high success rate in opening and traversing damped pull doors but also demonstrates strong robustness to external disturbances—capabilities that are difficult to realize with traditional methods.

% This work aims to leverage the powerful to solve everlasting challenge in mobile robots: Door opening and passing. 

% Door opening and traversal remain long-standing challenges in robotics, requiring precise coordination across perception, manipulation, and locomotion. Prior approaches have predominantly relied on model-based methods and state abstraction, often avoiding direct learning from raw RGB inputs. Some learning-based studies have demonstrated the possibility of door opening from RGB inputs, but they primarily focus on simplified scenarios that do not require the complex maneuvers involved in traversing a door.

% Recently, end-to-end learning directly from images has achieved remarkable success in robotic manipulation tasks requiring fine-grained control and long-horizon planning, through approaches such as diffusion policies and Transformer-based models. However, these advances have not yet been applied to the door traversal problem.

% In this work, we present, to the best of our knowledge, the first application of end-to-end learning from pixels to the door opening and traversal task, which demands tight coordination between a robot’s manipulator and its mobile base. Specifically, we demonstrate a learned policy capable of both opening a door and traversing through it. Our experiments show that training directly on raw visual observations is not only feasible but also effective for such integrated mobile manipulation tasks.

\end{abstract}

%%%%%%%%%%%%%%%%%%%%%%%%%%%%%%%%%%%%%%%%%%%%%%%%%%%%%%%%%%%%%%%%%%%%%%%%%%%%%%%%
\section{INTRODUCTION}

Door opening and passing with a dual-arm robot mounted on a nonholonomic mobile base presents unique challenges, including coordinating the arms with the base’s nonholonomic movements, performing constrained motions such as rotating and holding the doorknob, following the door’s circular trajectory, maintaining it in an open position, and counteracting self-closing mechanisms. Prior works often simplify the problem by removing these mechanisms, focusing only on opening without traversal, restricting to specific door types, or limiting interaction to pushing rather than pulling. In contrast, real human environments involve heavy, damped, self-closing doors that commonly require pulling. Humans naturally utilize their dual arms in a dexterous manner—rotating the knob, widening the gap, holding the door, switching arms as needed, and moving through while maintaining clearance. The contact-rich nature of such interactions highlights the gap between current simplifications and the complexities of real-world deployments.

Achieving high success rates in door opening and traversal, therefore, requires not only sophisticated motion planning but also tight coordination across perception, manipulation, and navigation, as well as robust recovery from failures caused by external disturbances. Traditional approaches rely on offline, model-based planning methods that assume prior knowledge of door geometry and dynamics \cite{5509445,chitta2010planning, stuede2019door, jang2023motion, sleiman2023versatile}, limiting adaptability and robustness in diverse environments. More recently, learning-based methods, including behavior cloning and reinforcement learning \cite{wang2020learning, Li2024UniDoorManip, khansari2023practical, Kang_2024, zhang2024learningopentraversedoors, xiong2024adaptive, HiroshiDoor}, have been applied to door-opening tasks. However, these frameworks often rely on abstracted state representations, require multiple sub-modules, or address simplified scenarios such as undamped push doors with minimal disturbances. In contrast, damped pull doors—common in buildings, workplaces, and public spaces—are far more representative of real-world settings. They introduce resistance that prevents simple pushing strategies, demand close coordination of arm and base movements, and require precise base trajectory control during traversal, while the self-closing mechanism acts as an inherent disturbance. Despite their practical importance, robustly opening and traversing damped pull doors using only visual input remains an unsolved problem.

% as it demands seamless loco-manipulation capability and robustness to unexpected door movement.
% as it demands not only effective manipulation but also close coordination between a robot’s arms and its mobile base.

Recent advances in diffusion-style policies have shown strong potential for end-to-end robotic learning, where visuomotor policies are trained directly from raw observations to actions. Unlike earlier methods that rely heavily on engineered state abstractions, diffusion-based approaches can capture complex dynamics and generate smooth, multi-step action sequences. Importantly, they have demonstrated the ability to learn effective manipulation and control behaviors from only a few demonstrations \cite{chi2023diffusion,zhao2023learning}, while also exhibiting strong robustness to external disturbances. These properties make diffusion-based policies a promising paradigm for complex real-world manipulation tasks. 
% such as door opening and traversal. 
However, prior research has primarily focused on fixed-base tabletop manipulation or navigation tasks. This raises a key question: can a diffusion policy handle long-horizon tasks like door opening and traversal that require tight coordination of dual arms with a nonholonomic base?

To answer this question, we created a dedicated framework for the door opening and traversal problem, including reference trajectory generation and diffusion policy training. For data generation, we designed a stage-based motion planner that integrates inverse kinematics (IK) for dual-arm manipulation with model predictive control (MPC) for nonholonomic base motion. A task-space scheduler was developed to provide a sequence of keyframes for IK, and based on these solutions, smooth joint trajectories were generated and executed by the robot in dynamic simulation, Mujoco \cite{6386109}. In addition, domain randomization of lighting and door/handle appearances was applied to improve robustness under scene variations. This setup enabled efficient data collection with minimal efforts.

For diffusion-based control policy, we adopted three independent ResNet-18 encoders for multi-view image features, concatenated with proprioceptive inputs, and a 1D U-Net with FiLM conditioning as the diffusion backbone. The policy was trained with a 16-step prediction horizon, 8-step execution before replanning, and 100 denoising steps, later reduced to as few as 10 at inference without loss of performance. As a result, we achieved a single end-to-end policy capable of executing the long-horizon action sequences to open a damped pull door and traverse through it.

We further validated the approach on real hardware using the RealMan dual-arm platform. For robot testing, we collected reference trajectories via a teleoperation kit with synchronized proprioceptive signals and multi-view RGB data. The same network architecture was employed during both training and inference. The learned policy successfully executed coordinated door-opening tasks and demonstrated robust recovery from external disturbances: when the door was re-closed during execution, the robot detected the change and re-initiated the opening sequence rather than failing or diverging. This resilience highlights the advantage of diffusion-based policies over traditional model-based methods and other learning-based approaches, which often lack recovery capabilities once disturbed.

In sum, our contributions are threefold: 1) a stage-based trajectory generation pipeline that combines inverse kinematics and model predictive control with randomized test scenarios, automating the generation of diverse reference trajectories, 2) the development of an image-to-action coordination controller based on diffusion policy, with a configuration that balances model complexity and inference time according to task completion performance, and 3) experimental validation on a real robot, demonstrating strong robustness through recovery behaviors using only vision and proprioceptive inputs. 
% To our knowledge, this is the first single policy that enables a mobile manipulator to successfully open and traverse a damped pull door.

\section{RELATED WORKS}

\subsection{Door Opening and Traversing}
Door opening and traversal have a long history in mobile manipulation. Early work predominantly adopted model-based formulations that simplify the door–robot interaction to make planning tractable. For example, \cite{chitta2010planning} generates whole-body motions under a fixed contact schedule while explicitly modeling the door as an articulated object. Jang et al. \cite{jang2023motion} decompose the problem into two stages—first planning over base pose and door angle, then solving arm motion via inverse kinematics—whereas \cite{sleiman2023versatile} enumerates a predefined set of contact options and combines graph search with trajectory optimization to achieve versatile loco-manipulation. However, during execution, most model-based methods rely on an auxiliary estimator to recover the door-handle pose (position and orientation). To improve autonomy, several works learn predictors for semantic keypoints such as door angle or handle pose \cite{wang2020learning,stuede2019door}; more recently, \cite{Wang2025DoorBot} leverages a vision–language model for handle detection and a learned grasp-offset predictor to improve grasp reliability. Nonetheless, most model-based pipelines rely on hand-engineered, discrete state machines that can be brittle outside their design assumptions.

Learning-based approaches have also been explored for door opening and traversal. Kang et al. and Zhang et al. \cite{Kang_2024,zhang2024learningopentraversedoors} train reinforcement learning policies to open and pass through doors, while Xiong et al. \cite{xiong2024adaptive} first learn with behavior cloning and then fine-tune on hardware using RL for door opening. However, these methods generally depend on auxiliary perception modules (e.g., handle detectors) \cite{urakami2019doorgym} and often target simplified settings. In contrast, Khansari et al. \cite{khansari2023practical} demonstrate a single neural network that performs door opening and traversal directly from vision and proprioception, but their evaluation focuses on undamped push doors, which do not require the regulation of the gap and tight arm-base coordination required by damped pull doors. \cite{HiroshiDoor} presented a learning-based approach for opening and passing through damped pull doors. However, due to the design of their network, the policy fails to generalize to doors with colors outside the training set and is also prone to failure when encountering disturbances.

\begin{figure*} [ht!]
    \centering
    \includegraphics[width=\linewidth]{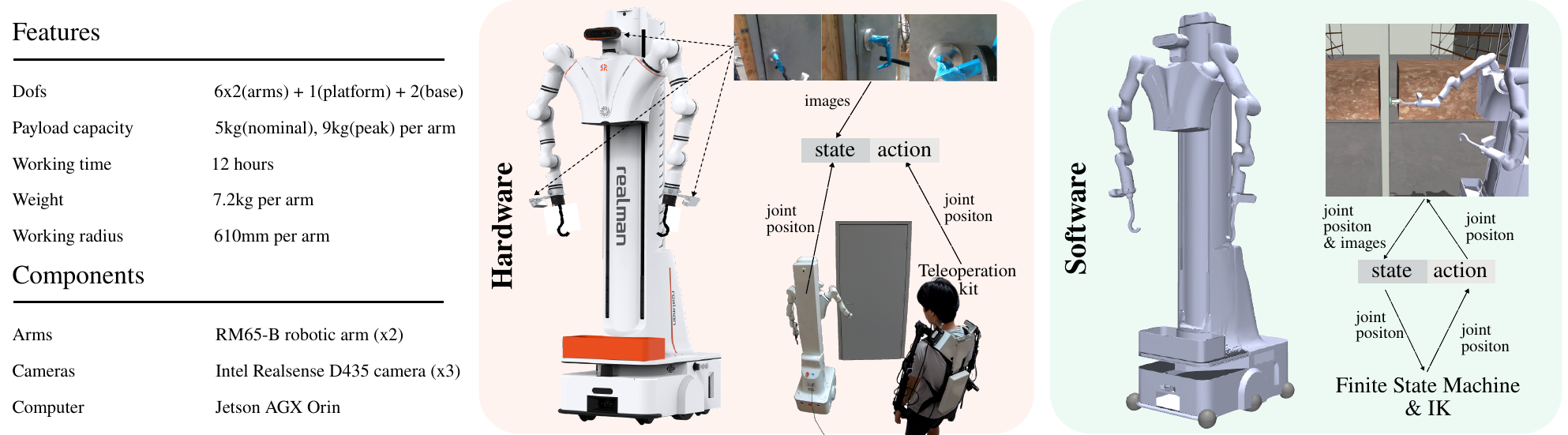}
    \caption{{\bf Hardware spec and data collection method.} Left: A RealMan platform equipped with two RM65-B robotic arms. Right: On hardware, a RealMan teleoperation kit is used for data collection, while in simulation, a state-based controller combining IK and MPC is employed.}  
    \label{fig:data_collection}
\end{figure*}
\subsection{Learning for Mobile Manipulation}
In recent years, learning-based methods have significantly advanced the field of mobile manipulation. However, navigation and manipulation are still often solved as separate subproblems. For example, \cite{pmlr-v164-sun22a,10160934,wu2023tidybot,gu2022multiskillmobilemanipulationobject} decompose long-horizon mobile manipulation into a repertoire of low-level visual skills, effectively decoupling manipulation from navigation. While such modular approaches can perform well in relatively simple tasks such as object fetching, they are insufficient for more complex scenarios like damped door traversal, which requires tight coordination between manipulation and locomotion. In these cases, a unified skill is essential for robust and safe execution.  

Recent algorithmic breakthroughs in imitation learning have made it possible to train end-to-end policies from only a handful of demonstrations \cite{chi2023diffusion,zhao2023learning}. For instance, \cite{Fu2024MobileAL} introduced a low-cost whole-body teleoperation system that demonstrated the ability to teach end-to-end mobile manipulation tasks with as few as 50 demonstrations.  These advances create a clear opportunity to develop an end-to-end policy that can open and traverse damped doors, where precise coordination between arms and base is indispensable.  

\section{METHOD}
Our objective is to learn an end-to-end door opening and traversal policy $\pi_\theta(a \mid s)$, parameterized by $\theta$, that maps an observation $s \in \mathcal{S}$ (RGB images and proprioceptive data) to a continuous action $a \in \mathcal{A}$. In the imitation learning setting, we assume access to a dataset of expert demonstrations
$$
\mathcal{D} = \{\tau^{*(i)}\}_{i=1}^N $$ 
$$ 
\tau^{*(i)} = (s^{(i)}_0, a^{(i)}_0, s^{(i)}_1, a^{(i)}_1, \dots, s^{(i)}_{T-1}, a^{(i)}_{T-1}, s^{(i)}_T),
$$
where each action $a^{(i)}_t$ is generated by an expert policy $\pi^{*}(a \mid s)$.

\begin{figure}[ht!]
    \centering
    \includegraphics[width=\linewidth]{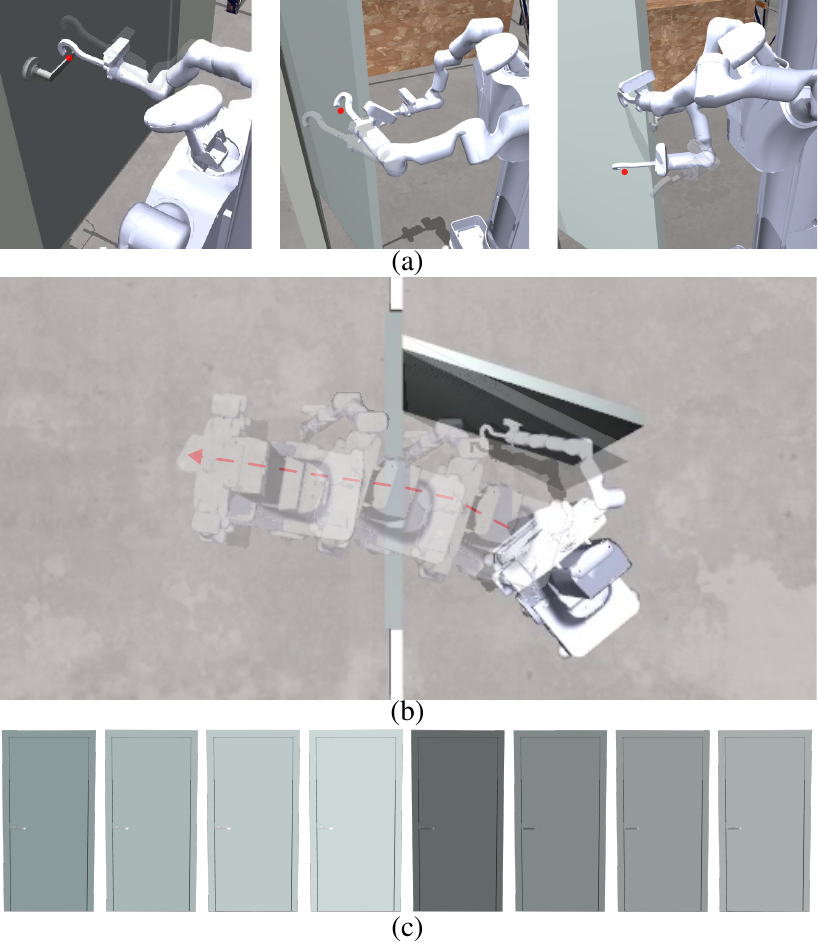}
    \caption{{\bf Data collection in simulation.} (a) To open the door, virtual keypoints are placed on the handle, and IK is used to compute the corresponding joint configurations. (b) Base locomotion is controlled by MPC, while the high-level state-based controller coordinates base movement with door manipulation. (c) To simulate visual variability present in the real world, door and handle appearances are randomized in each episode.}  
    \label{fig:simdatacollection}
\end{figure}

\subsection{Data Collection}

For collecting expert demonstrations on hardware, we use the teleoperation kit provided by Realman (Fig.~\ref{fig:data_collection}), which has the same joint configuration as the target robot. During teleoperation, the robot’s joint positions and images from all three cameras are recorded as the state, while the joint positions commanded through the teleoperation kit are stored as actions.

To collect data in simulation, we use a state-based controller that combines inverse kinematics (IK) for door manipulation with model predictive control (MPC) for base motion, as shown in Fig. \ref{fig:simdatacollection}. To better approximate human-like behavior, rather than directly performing joint-level control with the IK solution, we developed a scheduler that gradually guides the robot in the task-space toward the desired joint configuration found by the IK. To enhance realism and variability, we randomize lighting conditions as well as door and handle colors across episodes, thereby reflecting the natural diversity of real-world environments.

\begin{figure} [ht!]
    \centering
    \includegraphics[width=\linewidth]{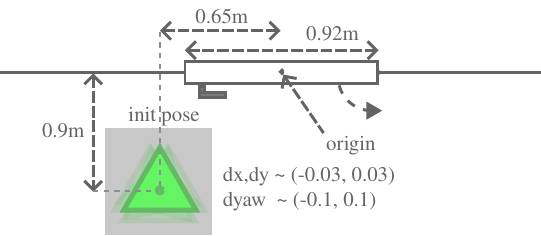}
    \caption{{\bf Task setting.} To increase the diversity of the training data, we randomize the robot’s initial base pose relative to the door. The robot base is place at a random lateral distance of $0.90\pm0.03$ with longitudinal offset of $\pm 0.03$m, and yaw of $\pm 1.00$ rad.}  
    \label{fig:init}
\end{figure}

In both hardware and simulation, we generated 100 demonstration trajectories. To prevent the policies from overfitting to a single motion pattern, we additionally randomize the robot’s initial base pose ($dx$, $dy$, and $d\text{yaw}$) as shown in Fig.~\ref{fig:init}.

\subsection{Observation and Action Space}
We collect image data from three independent cameras of shape $180\times240$ RGB images and use ResNet-18 to map each image to latent vectors $z^{(i)}_t|_{i=1}^{3} \in \mathbb{R}^{d_{\mathrm{cam}}}$. Concatenating the latent representation of images with state $s_t \in \mathbb{R}^{d_{\mathrm{state}}}$ yields
\[
o_t = \big[z^{(1)}_t, z^{(2)}_t, z^{(3)}_t, s_t\big] \in \mathbb{R}^{d_o}, 
\quad d_o = 3d_{\mathrm{cam}} + d_{\mathrm{state}}.
\]
Our policy conditions on the most recent $T_o$ observations
\[
O_t = [\,o_{t-T_o+1},\,\dots,\,o_t\,] \in \mathbb{R}^{T_o \times d_o},
\]
and predicts a length-$T_p$ future action sequence
\[
A_t = [\,a_t,\,a_{t+1},\,\dots,\,a_{t+T_p-1}\,] \in \mathbb{R}^{T_p \times d_a}.
\]
During inference, we perform only the first $T_a$ steps where $T_a \le T_p$.

% \subsection{Conditional Density}
% We model the conditional density of future actions given recent observations:
% $$
% p_\theta(A_t \mid O_t).
% $$
% Unlike DDPM, in which we model the joint $p(A_t, O_t)$, conditioning on $O_t$ allows us to focus on modeling $A_t$, reducing variance and allowing faster denoising suitable for real time control \cite{chi2023diffusion}.

\begin{figure} 
    \centering
    \includegraphics[width=\linewidth]{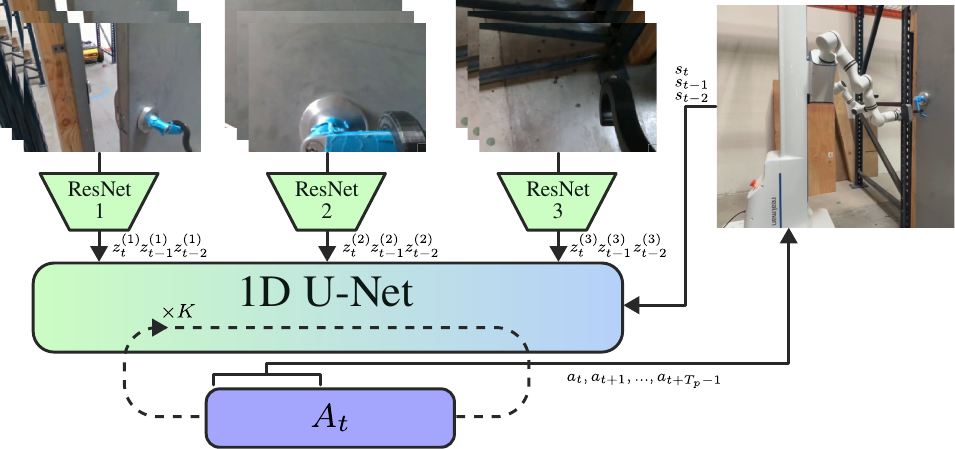}
    \caption{{\bf Diffusion Policy.} We model $p_\theta(A_t|O_t)$ with three ResNet-18 Visual encoder for each camera and 1D U-Net with FiLM conditioning. During inference, we perform $K$ denoising steps to transform gaussian noise into an action sequence $A_t$.} 
    \label{fig:network_arch}
\end{figure}

\subsection{Diffusion Policy Inference and Training}
At each denoising period of inference, the diffusion policy ingests the latest $O_t$ and predicts $A_t$ (Fig.~\ref{fig:network_arch}). We sample $ A^{k}_{t} \sim \mathcal{N} (\boldsymbol{0}, \sigma^2I)$ and perform $K$ denoising steps:

$$
A^{k-1}_{t} \leftarrow \alpha(A^{k}_{t} - \gamma \epsilon_{\theta}(O_t,A^{k}_{t},k) + \mathcal{N}(0, \sigma^2I))
$$

By performing the denoising steps $K$ times, we obtain $A^{k}_{t}, A^{k-1}_{t}, ...$, until we obtain our desired denoised action horizon $A^{0}_{t}$ \cite{chi2023diffusion}. To train our noise prediction network $\epsilon_\theta$, we approximate $\arg\min_\theta \mathcal{L(\theta)}$ via gradient descent, where:
$$\mathcal{L}(\theta) = MSE(\epsilon^k,\epsilon_\theta(O_t, A_t^0+\epsilon^k, k))$$

\subsection{Hyperparameter Selection}

\begin{table}[h]
\caption{Hyperparameters}
\label{table_example}
\begin{center}
\begin{tabular}{|c||c|}
\hline
\textbf{Hyperparameter} & \textbf{Value} \\
\hline
Vision Backbone & ResNet-18\\
\hline
Prediction Horizon $T_p$ & 16 \\
\hline
Action Sequence $A_t$ & 8\\
\hline
Stacked Observation Size $T_o $& 3\\ 
\hline
Forward Diffusion steps  $K$ & 100\\
\hline
\end{tabular}
\end{center}
\end{table}

We found that careful selection of hyperparameters is crucial for policy success (Table~\ref{table_example}.). To capture short-term motion cues while avoiding excessive memory overhead, we stack the three most recent states ($T_o=3$) as input observations. Each camera stream is processed by an independent pretrained ResNet-18 encoder to extract view-specific features. The prediction horizon ($T_p=16$) provides sufficient look-ahead to execute sub-skills while limiting error accumulation, and the policy executes $A_t=8$ actions before replanning to remain responsive to new observations. For the diffusion process, we use $K=100$ forward steps during training to ensure a smooth noise schedule and stable optimization, while at inference we employ fewer denoising steps (see Sec.~IV) to reduce latency and enable real-time control. The effect of different hyperparameter choices on performance is further analyzed in the following section.

\section{EXPERIMENT}

\begin{figure*} 
    \centering
    \includegraphics[width=0.95\linewidth]{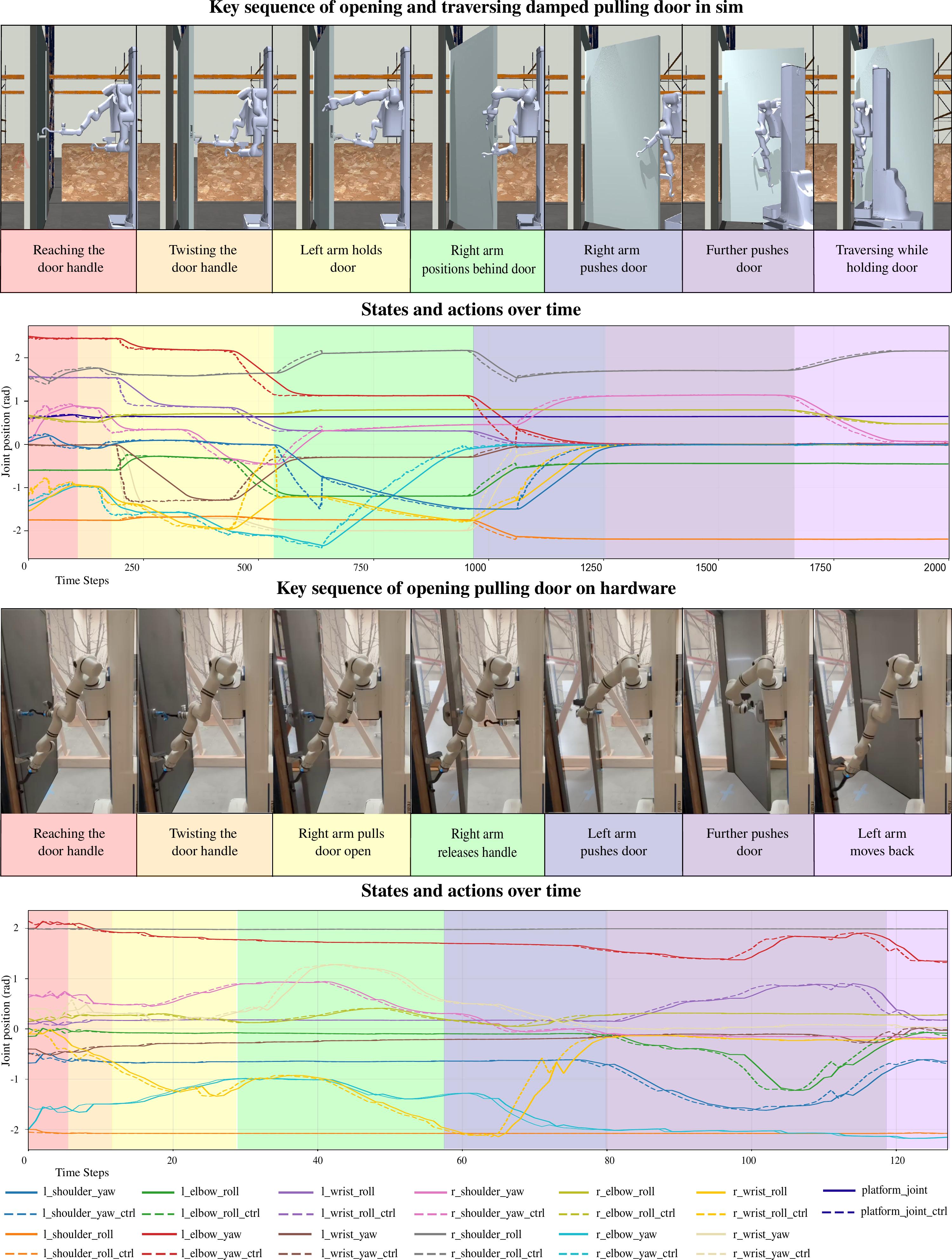}
    \caption{{\bf Policy deployment.} Rollout of the trained policy with synchronized state and action in sim and hardware. Solid line denotes state $s_t$ and dotted line denotes action/control $a_t$. Colored band denotes key behavioral stages, matched to color coded robot images and labels.}  
    \label{fig:experiment}
\end{figure*}

\begin{figure*}[ht!]
    \centering
    \includegraphics[width=\linewidth]{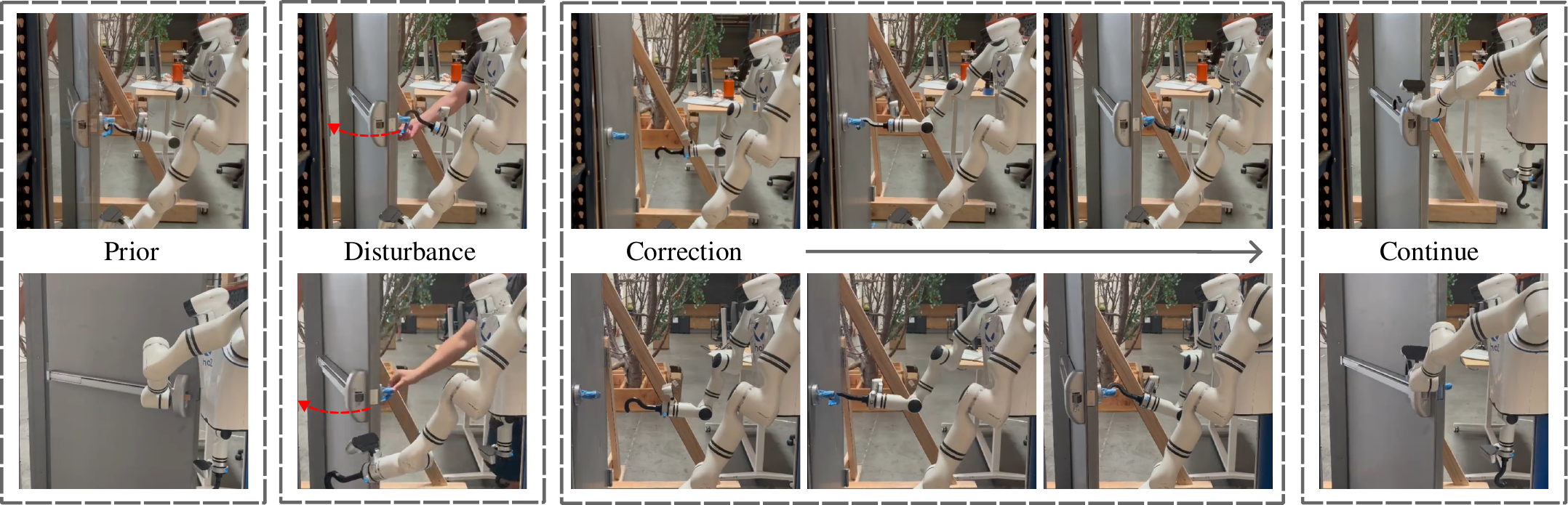}
    \caption{{\bf Policy under external disturbance during door opening.} (Left) Prior: the robot initially grasps and begins to pull the door. (Second) Disturbance: the door is manually re-closed while the robot is executing the opening motion. (Middle) Correction: the policy reacts by halting the current arm extension, re-adjusting its posture, and re-initiating the opening sequence. (Right) Continue: the robot successfully resumes and completes the door-opening task.}  
    \label{fig:disturbance}
\end{figure*}

\subsection{Platform selection}
We select the RealMan double arm lifting platform as the hardware platform. As shown on the left of Fig.~\ref{fig:data_collection}, the robot has two 6-DoF arms with a maximum 5 kg payload per arm, a platform joint which is able to change the torso position, and a non-holonomic mobile base.

\subsection{Simulation Results and Ablation Study}

We successfully trained a diffusion policy that enables a mobile manipulator to open and traverse a damped pull door using only visual and proprioceptive inputs. Accomplishing this task requires the policy to learn a long-horizon trajectory that integrates multiple coordinated skills, including reaching for the handle, twisting it, pulling the door, coordinating both arms, and synchronizing manipulation with locomotion. (see Fig.~\ref{fig:experiment}).

We conducted an ablation study on our proposed policy and additionally trained ACT \cite{zhao2023learning} and SmolVLA \cite{shukor2025} as baselines for comparison as shown in Table.~\ref{table:ablation_study}. Each policy was trained for 100k steps, with evaluations performed every 20k steps. During each evaluation, the policy was executed 10 times, and we report the best-performing trial. This evaluation strategy was adopted because, in imitation learning, low training loss does not necessarily lead to high task success rate.

Our results reveal several key insights. First, using a unified visual backbone that shares weights across all three cameras consistently reduced task success compared to view-specific encoders. We hypothesize that the shared backbone suffers from multi-view domain shift: each camera provides inputs with distinct geometry, scale, and statistical properties, resulting in diluted representations that fail to capture view-specific features. This limitation is especially detrimental in tasks requiring fine-grained perception, such as precisely locating and manipulating the door handle. Additionally, we found that without visual randomization (e.g., color and lighting), the policy was highly brittle, often failing when even minor changes such as door color were introduced. In contrast, with randomization applied during training, the policy demonstrated strong robustness and was able to generalize successfully across diverse lighting conditions and door appearances.

Second, we evaluated the effect of reducing the number of denoising steps in the diffusion process during training. We found that performance remained stable with as few as 30 steps, while the inference time was reduced substantially. Performance began to degrade when the number of steps was decreased further. Interestingly, during deployment we observed that the policy remained effective even when the number of denoising steps was reduced to as few as 10, despite being trained with a much larger number of steps. This significantly improved inference speed without sacrificing task success rate.

Third, we observed that ACT converged significantly faster than our diffusion-based policy. However, it could not achieve a 100\% success rate. This gap may be due to the domain randomization introduced in the simulation, such as variations in door and handle color. Transformer-based policies like ACT tend to overfit to dominant visual patterns and often struggle to capture multi-modal behaviors, which makes them less robust under randomized conditions. In contrast, diffusion policies explicitly model action distributions through iterative denoising, allowing them to better handle visual variability and recover from disturbances. We also tested SmolVLA, but the model consistently failed to complete the task even after 100k training steps. We suspect this limitation arises because SmolVLA is pretrained primarily on community-collected datasets involving smaller manipulators, which reduces its transferability to our larger, more complex setting.

\begin{table}[h]
\caption{Simulation Result}
\label{table:ablation_study}
\begin{center}
\begin{tabular}{|c||c|}
\hline
Policy & Success rate\\
\hline
Diffusion (unified encoder denoising step 100) & 0/10\\
\hline
Diffusion (separate encoder denoising step 100) & 10/10 \\
\hline
Diffusion (separate encoder denoising step 30)& 10/10 \\ 
\hline
Diffusion (separate encoder denoising step 20)& 9/10\\
\hline
ACT & 8/10\\
\hline
SmolVLA & 0/10 \\
\hline
\end{tabular}
\end{center}
\end{table}

\subsection{Hardware Results}

% To further assess the effectiveness of our approach, we deployed the policy on real hardware. The policy successfully executed reliable door-opening behaviors with bimanual coordination (Fig.~\ref{fig:experiment}). Notably, the learned policy demonstrated robustness to disturbances (Fig.~\ref{fig:disturbance}). For example, after the robot partially opened the door, we manually re-closed it. The policy responded by halting further extension with the left arm and instead re-initiating the opening sequence. Similarly, when the door was re-closed after being fully extended by the right arm, the policy detected the change and restarted the opening motion. These results indicate that the policy relies on visual feedback to adapt its behavior, an essential property for deployment in real-world environments. To deploy
% Due to limited onboard computational resources, however, the relatively high latency of diffusion-based inference (stemming from the denoising steps) prevented integration of base control, since the base is driven by velocity commands that require low-latency responses. Nevertheless, the robot was able to learn and execute a complex opening sequence, and with appropriate hardware acceleration, full door traversal can be readily achieved.

To further evaluate the effectiveness of our approach, we deployed the learned policy on real hardware. The policy successfully executed reliable door-opening behaviors with coordinated bimanual manipulation (Fig.~\ref{fig:experiment}). Notably, the learned policy demonstrated robustness to disturbances (Fig.~\ref{fig:disturbance}).

For example, after the robot partially opened the door, we manually pushed it closed. In response, the policy halted further extension of the left arm and re-initiated the opening sequence. Similarly, when the door was manually closed after being fully extended by the right arm, the policy detected the state change and restarted the opening motion. These behaviors indicate that the policy actively relies on visual feedback to adapt to environmental changes—an essential capability for deployment in real-world settings.

To achieve full door traversal, low-latency base control is essential. To meet this requirement, we incorporated LoRA (Low-Rank Adaptation) with quantization to reduce inference time and improve computational efficiency. This significantly lowered policy latency, enabling faster feedback between perception and control. In addition, we optimized the onboard computer’s power configuration to unlock higher sustained compute performance. By adjusting the system power mode, we reduced frequency throttling and stabilized real-time inference throughput. The combination of model-level optimization and hardware-level tuning ensured consistent low-latency execution, allowing the policy to successfully complete full door traversal.

\section{Conclusion}

Our study demonstrates that diffusion-based visuomotor policies can achieve reliable performance on the challenging task of opening and traversing damped pull doors using a dual-arm mobile manipulator. Unlike prior approaches that rely on state machines or heavily engineered perception pipelines, our method learns a unified policy that integrates perception, manipulation, and base coordination directly from demonstration data. The results show that diffusion policies not only generate long-horizon trajectories but also exhibit robustness to disturbances and environmental variability, a critical capability for real-world deployment.


%%%%%%%%%%%%%%%%%%%%%%%%%%%%%%%%%%%%%%%%%%%%%%%%%%%%%%%%%%%%%%%%%%%%%%%%%%%%%%%%

%%%%%%%%%%%%%%%%%%%%%%%%%%%%%%%%%%%%%%%%%%%%%%%%%%%%%%%%%%%%%%%%%%%%%%%%%%%%%%%%

%%%%%%%%%%%%%%%%%%%%%%%%%%%%%%%%%%%%%%%%%%%%%%%%%%%%%%%%%%%%%%%%%%%%%%%%%%%%%%%%

\bibliographystyle{IEEEtran}
\bibliography{references}

% \begin{thebibliography}{99}

% \bibitem{c1} 
% \bibitem{c2} Y. Karayiannidis, C. Smith, F. E. Vi˜na, P. Ogren, and D. Kragic, ""Open Sesame!" Adaptive Force/Velocity Control for Opening Unknown Doors," in *Proc. IEEE/RSJ Int. Conf. Intell, Robots Syst. (IROS)*, Vilamoura, Algarve, Portugal, Oct. 2012, pp. 4040-4047
% \bibitem{c3} Z. Wang, Y. Mo, S. Jin, and W. Yuan, "DoorBot: Closed-Loop Task Planning and Manipulation for Door Opening in the Wild with Haptic Feedback," arXiv preprint arXiv:2504.09358, Apr. 2025. doi: 10.458550/arXiv.2504.09358.
% \bibitem{c4} M. Stuede, K. Nuelle, S. Tappe, and T. Ortmaier, "Door opening and traversal with an industrial cartesian impedance controlled mobile robot," in *Proc. IEEE Int. Robot. Autom. (ICRA)*, 2019, pp. 966-972, doi: 10.1109/ICRA.2019.8793866.

% \end{thebibliography}

\end{document}